\documentclass[11pt,a4paper]{article}
\usepackage[hyperref]{naaclhlt2019}
\usepackage{times}
\usepackage{latexsym}

\usepackage{url}
\usepackage{soul}

\usepackage{amsmath,graphicx}
\usepackage{multirow,multicol}
\usepackage{amssymb}
\usepackage{booktabs}
\usepackage{enumitem}

\aclfinalcopy

\usepackage{tikz}
\usetikzlibrary{calc,arrows.meta}
\usetikzlibrary{positioning}
\newdimen\XCoord
\newdimen\YCoord

\newcommand*{\ExtractCoordinate}[1]{\path (#1); \pgfgetlastxy{\XCoord}{\YCoord};}%

\newcommand{\youcook}[0]{\texttt{YouCook2}}
\newcommand{\smg}[0]{\texttt{sth-sth}}

\usepackage{manyfoot}
\SetFootnoteHook{\hspace*{-1.8em}}
\DeclareNewFootnote{bl}[gobble]
\title{Neural Language Modeling with Visual Features}

\author{Antonios Anastasopoulos$^{1,\ast}$ 
Shankar Kumar$^2$ and Hank Liao$^2$ \\
  $^1$Language Technologies Institute, Carnegie Mellon University \\
  $^2$Google Inc., NY\\
  {\tt aanastas@cs.cmu.edu, \{shankarkumar,hankliao\}@google.com}
}

\date{}

\begin{document}
%
\maketitle
\begin{abstract}
Multimodal language models attempt to incorporate non-linguistic features for the language modeling task. In this work, we extend a standard recurrent neural network (RNN) language model with features derived from videos. We train our models on data that is two orders-of-magnitude bigger than datasets used in prior work. We perform a thorough exploration of model architectures for combining visual and text features. Our experiments on two corpora (\textit{YouCookII} and \textit{20bn-something-something-v2}) show that the best performing architecture consists of middle fusion of visual and text features, yielding over $25\%$ relative improvement in perplexity. We report analysis that provides insights into why our multimodal language model improves upon a standard RNN language model.
\end{abstract}

\section{Introduction}
\label{sec:intro}

\footnotebl{$^\ast$Work performed while the author was an intern at Google.}

Language models are vital components of a wide variety of systems for Natural Language Processing (NLP) including Automatic Speech Recognition, Machine Translation, Optical Character Recognition, Spelling Correction, etc. However, most language models are trained and applied in a manner that is oblivious to the environment in which human language operates~\cite{ororbia2018visually}.
These models are typically trained only on sequences of words, ignoring the physical context in which the symbolic representations are grounded, or ignoring the social context that could inform the semantics of an utterance.

For incorporating additional modalities, the NLP community has typically used datasets such as MS COCO \cite{lin2014microsoft} and Flickr \cite{rashtchian2010collecting} for image-based tasks, while several datasets \cite{chen2011collecting,yeung2014videoset,das2013thousand,rohrbach2013translating,hendricks2017localizing} have been curated for video-based tasks. Despite the lack of big datasets, researchers have started investigating language grounding in images \cite{plummer2015flickr30k,rohrbach2016grounding,socher2014grounded} and to lesser extent in videos \cite{regneri2013grounding,lin2014microsoft}. However, language grounding has focused more on obtaining better word and sentence representations or other downstream tasks, and to lesser extent on language modeling.

In this paper, we examine the problem of incorporating temporal visual context into a recurrent neural language model (RNNLM). Multimodal Neural Language Models were introduced in \cite{kiros2014multimodal}, where log-linear LMs \cite{mnih2007three} were conditioned to handle both image and text modalities. Notably, this work did not use the recurrent neural model paradigm which has now become the \emph{de facto} way of implementing neural LMs. 

The closest work to ours is that of \citet{ororbia2018visually}, who report perplexity gains of around 5--6\% on three languages on the MS COCO dataset (with an English vocabulary of only 16K words).

\tikzset{vemb/.pic={%
\draw[rounded corners] (-5pt,0pt) rectangle (5pt,30pt);
\fill[color=red] (0pt,5pt) circle[radius=4pt]
                    (0pt,15pt) circle[radius=4pt]
                    (0pt,25pt) circle[radius=4pt];
}}
\tikzset{wemb/.pic={%
\draw[rounded corners] (-5pt,0pt) rectangle (5pt,30pt);
\fill[color=blue] (0pt,5pt) circle[radius=4pt]
                    (0pt,15pt) circle[radius=4pt]
                    (0pt,25pt) circle[radius=4pt];
}}

\tikzset{
  pics/emb/.style n args={4}{
    code = { %
        \node (#4) at (#2,#3) [draw,rounded corners,minimum width=8pt,minimum height=25pt] {};
        \foreach \i in {-0.25, 0,0.25}{
            \pgfmathtruncatemacro{\j}{\i + #3}
            \fill[color=#1] (#2,{\i + #3}) circle[radius=2.5pt];
        }
    }
  }
}

\tikzset{
  pics/cell/.style n args={4}{
    code = { %
        \node (#4) at (#2,#3) [draw,rounded corners,font=\tiny] {#1};
    }
  }
}

\begin{figure*}[h!]
    \centering
\begin{tikzpicture}
\node [outer sep=0] (A) at (0,0) {$<$S$>$ We will make muffins};
\pic {emb={blue}{-1.8}{1.2}{w1}};
\pic {emb={blue}{-0.9}{1.2}{w2}};
\pic {emb={blue}{0}{1.2}{w3}};
\pic {emb={blue}{0.9}{1.2}{w4}};
\pic {emb={blue}{1.8}{1.2}{w5}};

\draw [-latex] ($(A.north west)!.11!(A.north east)$) -- (w1.south);
\draw [-latex] ($(A.north west)!.28!(A.north east)$) -- (w2.south);
\draw [-latex] ($(A.north west)!.44!(A.north east)$) -- (w3.south);
\draw [-latex] ($(A.north west)!.63!(A.north east)$) -- (w4.south);
\draw [-latex] ($(A.north west)!.85!(A.north east)$) -- (w5.south);

\node [inner sep=0,outer sep=0] (B) at (5,0) {\includegraphics[scale=.075]{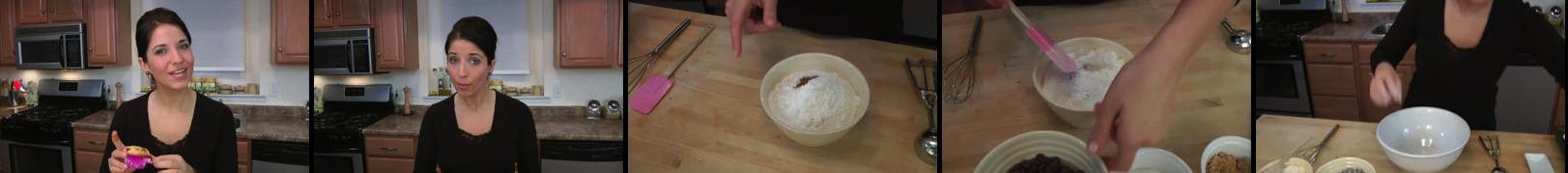}};

\ExtractCoordinate{$(B.north west)!.1!(B.north east)$};
\pic {emb={red}{\XCoord}{1.2}{v1}};
\ExtractCoordinate{$(B.north west)!.3!(B.north east)$};
\pic {emb={red}{\XCoord}{1.2}{v2}};
\ExtractCoordinate{$(B.north west)!.5!(B.north east)$};
\pic {emb={red}{\XCoord}{1.2}{v3}};
\ExtractCoordinate{$(B.north west)!.7!(B.north east)$};
\pic {emb={red}{\XCoord}{1.2}{v4}};
\ExtractCoordinate{$(B.north west)!.9!(B.north east)$};
\pic {emb={red}{\XCoord}{1.2}{v5}};

\draw[-latex] ($(B.north west)!.1!(B.north east)$) -- (v1.south);
\draw[-latex] ($(B.north west)!.3!(B.north east)$) -- (v2.south);
\draw[-latex] ($(B.north west)!.5!(B.north east)$) -- (v3.south);
\draw[-latex] ($(B.north west)!.7!(B.north east)$) -- (v4.south);
\draw[-latex] ($(B.north west)!.9!(B.north east)$) -- (v5.south);

\pic {cell={lstm}{-1.8}{3.2}{c11}};
\node (oplus1) [outer sep=0,inner sep=0,below=.85cm  of c11] {\tiny{$\oplus$}};
\draw[-latex] (w1.north) -- (oplus1.south);
\draw[-latex] (v1.north) |- (oplus1.east);
\draw[-latex] (oplus1.north) -- (c11.south);

\pic {cell={lstm}{-1.8}{4}{c21}};
\draw[-latex] (c11.north) -- (c21.south);

\node (o1) [above=0.4cm of c21] {\small{We}};
\draw[-latex] (c21.north) -- (o1.south);

\pic {cell={lstm}{-0.9}{3.2}{c12}};
\node (oplus2) [outer sep=0,inner sep=0,below=.6cm  of c12] {\tiny{$\oplus$}};
\draw[-latex] (w2.north) -- (oplus2.south);
\draw[-latex] (v2.north) |- (oplus2.east);
\draw[-latex] (oplus2.north) -- (c12.south);

\pic {cell={lstm}{-0.9}{4}{c22}};
\draw[-latex] (c12.north) -- (c22.south);

\node (o2) [above=0.4cm of c22] {\small{will}};
\draw[-latex] (c22.north) -- (o2.south);

\draw[-latex] (c11.east) -- (c12.west);
\draw[-latex] (c21.east) -- (c22.west);

\node (d1) [right=0.1 of c22] {...};
\node (d2) [right=0.1 of c12] {...};
\node (d3) [above=0.45 of d1] {...};

\node (l1) [above=0.3 of o2] {\textbf{Early Fusion}};

\draw [line width=0.4mm] (.7,3) -- (.7,6);

\pic {cell={lstm}{2.2}{2.8}{c211}};
\node (oplus3) [outer sep=0,inner sep=0,above=.35cm  of c211] {\tiny{$\oplus$}};
\draw[-latex,color=gray] (w3.north) -- +(0,0.3) -| (c211.south);
\draw[-latex,color=gray] (v3.north) -- +(0,0.8) -- +(-1.8,0.8) |- (oplus3.east);
\draw[-latex] (c211.north) -- (oplus3.south);

\pic {cell={lstm}{2.2}{4}{c221}};
\draw[-latex] (oplus3.north) -- (c221.south);

\node (o21) [above=0.4cm of c221] {\small{make}};
\draw[-latex] (c221.north) -- (o21.south);

\node (d21) [right=0.1 of c221] {...};
\node (d22) [right=0.1 of c211] {...};
\node (d23) [above=0.45 of d21] {...};
\node (d21a) [left=0.1 of c221] {...};
\node (d22a) [left=0.1 of c211] {...};
\node (d23a) [above=0.45 of d21a] {...};

\node (l2) [right=.8 of l1] {\textbf{Middle Fusion}};

\draw [line width=0.4mm] (3.8,3) -- (3.8,6);

\pic {cell={lstm}{5.5}{2.8}{c311}};
\pic {cell={lstm}{5.5}{3.6}{c321}};
\node (oplus5) [outer sep=0,inner sep=0,above=.35cm  of c321] {\tiny{$\oplus$}};
\node (o22) [above=0.35cm of oplus5] {\small{muffins}};

\draw[-latex,line width=0.02mm] (w4.north) -- +(0,0.15) -| (c311.south);
\draw[-latex] (c311.north) -- (c321.south);
\draw[-latex] (c321.north) -- (oplus5.south);
\draw[-latex,line width=0.02mm] (v4.north) |- (oplus5.east);
\draw[-latex] (oplus5.north) -- (o22.south);

\node (d31) [right=0.15 of c311] {...};
\node (d32) [right=0.15 of c321] {...};
\node (d33) [above=0.9 of d32] {...};
\node (d31a) [left=0.15 of c311] {...};
\node (d32a) [left=0.15 of c321] {...};
\node (d33a) [above=0.9 of d32a] {...};

\node (l3) [right=1.0cm of l2] {\textbf{Late Fusion}};

\draw [line width=0.4mm] (7.3,.4) -- (7.3,6);

\node (l4) [right=1.4 of l3] {\textbf{Linear Combination}};
\pic {emb={blue}{8.5}{4.7}{wa}};
\pic {emb={red}{8.5}{3.7}{va}};
\node (t1) [outer sep=0,inner sep=0,right=0.5 of wa] {\tiny{$\otimes$}};
\node (t2) [outer sep=0,inner sep=0,right=0.5 of va] {\tiny{$\otimes$}};
\node (t3) [draw,outer sep=0,above=0.25 of t2] {\tiny{add}};

\node (ww) [outer sep=0,right=0.4 of t1] {\tiny{$W^w$}};
\node (wv) [outer sep=0,right=0.4 of t2] {\tiny{$W^v$}};

\node (t4) [outer sep=0,right=0.6 of t3] {\tiny{input to next layer}};

\draw[-latex] (wa.east) -- (t1.west);
\draw[-latex] (va.east) -- (t2.west);
\draw[-latex] (ww.west) -- (t1.east);
\draw[-latex] (wv.west) -- (t2.east);

\draw[-latex] (t1.south) -- (t3.north);
\draw[-latex] (t2.north) -- (t3.south);
\draw[-latex] (t3.east) -- (t4.west);

\draw [line width=0.4mm, dashed] (7.4,3) -- (12.5,3);

\node (l4) [below=2.5 of l4] {\textbf{Weighting}};
\pic {emb={blue}{8.5}{2}{wb}};
\pic {emb={red}{8.5}{1}{vb}};
\node (t1b) [outer sep=0,inner sep=0,below right=0 and 0.3 of wb] {\tiny{$\odot$}};
\node (t2b) [draw,outer sep=0,right=0.3 of t1b] {\tiny{$\sigma$}};
\node (t3b) [draw,outer sep=0,right=0.3 of t2b] {\tiny{mul}};
\node (t4b) [outer sep=0,inner sep=0, above=0.25 of t3b] {\tiny{$\oplus$}};
\node (t5b) [outer sep=0,right=0.4 of t4b] {\tiny{input to next layer}};

\draw[-latex] (wb.east) -| (t1b.north);
\draw[-latex] (wb.east) -- (t4b.west);
\draw[-latex] (vb.east) -| (t1b.south);
\draw[-latex] (vb.east) -| (t3b.south);

\draw[-latex] (t1b.east) -- (t2b.west);
\draw[-latex] (t2b.east) -- (t3b.west);
\draw[-latex] (t3b.north) -- (t4b.south);
\draw[-latex] (t4b.east) -- (t5b.west);

\end{tikzpicture}

    \caption{Visualization of our different Language Models. Given word and visual embeddings, the input can be created by three methods. Left panels: simple concatenation (examples with early, middle, and late fusion of the visual embeddings). Top right panel: learning a linear combination of the two embeddings. Bottom right panel: learn to weight the visual embedding based on the current word. \textit{Note:} $\oplus$ denotes concatenation, $\otimes$ denotes matrix multiplication, $\odot$ denotes dot product.}
    \label{fig:models}
\end{figure*}
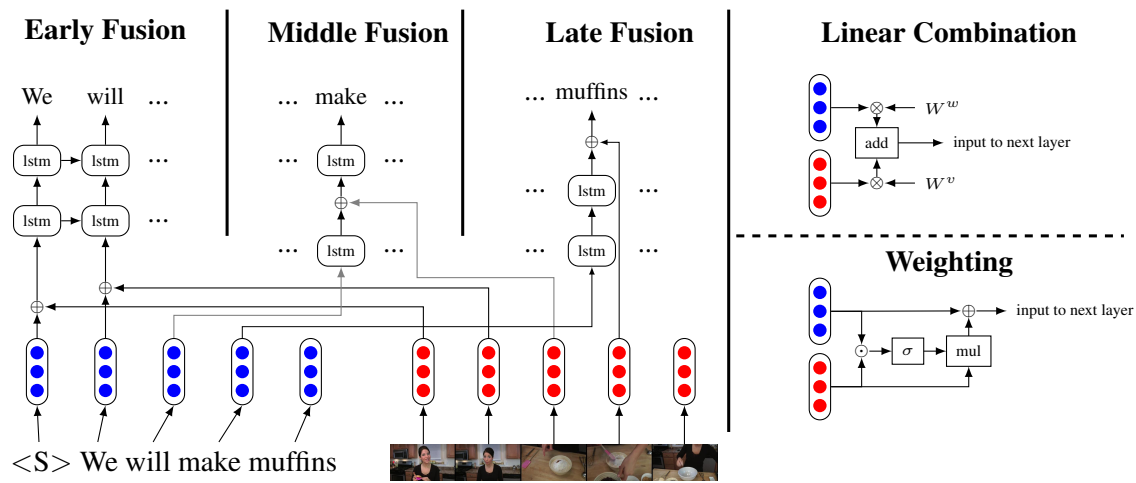

Our work is distinguishable from previous work with respect to three dimensions:
\begin{enumerate}[noitemsep,nolistsep]
    \item We train our model on video transcriptions comprised of text and visual features. Thus, both modalities of our model are temporal, in contrast to most previous work which uses static images. At the same time, our model respects the temporal alignment between the two modalities, combining the text with its concurrent visual context, mimicking a real natural language understanding situation.
    \item We explore several architectures for combining the two modalities, and our best model reduces perplexity by more than 25\% relative to a text-only baseline.
    \item The scale of our experiments is unprecedented: we train our models on two orders of magnitude more data than any previous work. This results in quite strong, hard-to-beat baselines.
\end{enumerate}

\section{Model}
\label{sec:model}
A language model assigns to a sentence $W=w_1\ldots w_M$ the probability:
\[
p(W) = \prod_{m=1}^M p(w_m \mid w_{<m})
\]
where each word is assigned a probability given the previous word history.

For a given video segment, we assume that there is a sequence of $N$ video frames represented by features $V=v_1\ldots v_N$, and the corresponding transcription $W=w_1\ldots w_M$. In practice, we assume $N=M$ since we can always assign a video frame to each word by replicating the video frames the requisite number of times. Thus, our visually-grounded language model models the probability of the next word given the history of previous words as well as video frames:
\[
p(W) = \prod_{m=1}^M p(w_m \mid w_{<m} , v_{<m})
\]

\subsection{Combining the text and video modalities}
There are several options for combining the text and video modalities. We opt for the simplest strategy, which concatenates the representations.
For a word embedding $w_i$ and corresponding visual representation $v_i$, the input to our RNNLM will be the concatenated vector $e_i = [w_i \ ; v_i]$. 
For the examples where we were unable to compute visual features (see Section \S\ref{sec:data}), we set $v_i$ to be a zero-vector.

In addition to concatenating the word and visual embedding, we explore two variants of our model that allow for a finer-grained integration of the two modalities:

\paragraph{a. Learning a linear combination of the two modalities} In this case, the RNNLM is given as input a vector $e_i$ that is a weighted sum of the two embeddings:
\[
    e_i = K^ww_i + K^vv_i
\]
where $K^w, K^v$ are learned matrices.
\paragraph{b. Weighting the visual embedding conditioned on the word} Here, we apply the intuition that some words could provide information as to whether or not the visual context is helpful. In a simplistic example, if the word history is the article ``the," then the visual context could provide relevant information needed for predicting the next word. For other word histories, though, the visual context might not be needed or be even irrelevant for the next word prediction: if the previous word is ``carpe", the next word is very likely to be ``diem", regardless of visual context.
We implement a simple weighting mechanism that learns a scalar weight for the visual embedding prior to concatenation with the word embedding. The input to the RNNLM is now $e_i = [w_i \ ; \lambda v_i]$, where:
    \[
    \lambda = \sigma(w_i \cdot v_i).
    \]
This approach does not add any new parameters to the model, but since the word representations $w_i$ are learned, this mechanism has the potential to learn word embeddings that are also appropriate for weighting the visual context.

\subsection{Location of combination}
We explore three locations for fusing visual features in an RNNLM (Figure~\ref{fig:models}). Our \textit{Early Fusion} strategy merges the text and the visual features at the input to the LSTM cells. This embodies the intuition that it is best to do feature combination at the earliest possible stage. The \textit{Middle Fusion} merges the visual features at the output of the $1^{\text{st}}$ LSTM layer while the \textit{Late Fusion} strategies merges the two features after the final LSTM layer. The idea behind the \textit{Middle} and \textit{Late} fusion is that we would like to minimize changes to the regular RNNLM architecture at the early stages and still be able to benefit from the visual features.  

\begin{table}[t]
    \centering
    \begin{tabular}{c|c|c}
    \toprule
        \multirow{2}{*}{Model} & \multicolumn{2}{c}{Perplexity (Reduction)}\\
        &  \youcook & \smg \\
    \midrule
        text-only & 89.8  & 513.6\\
    \midrule
        Linear Comb. & 84.8 (6\%) & 580.8 (--)\\
        Weighting & 76.8 (14\%) & 538.8 (--) \\
        Early Fusion & 93.7 (--) & 611.3 (--) \\
        Middle Fusion & \textbf{64.9 (28\%)}  & \textbf{411.4 (20\%)} \\
        Late Fusion & 79.3 (12\%) & 485.5 (5\%) \\
    \bottomrule
    \end{tabular}
    \caption{\textit{Middle Fusion} of text and frame-level visual features leads to significant reductions in perplexity on two multimodal datasets.}
    \label{tab:ppl}
\end{table}
\section{Data and Experimental Setup}
\label{sec:data}
Our training data consist of about $64$M segments from YouTube videos comprising a total of $1.2$B tokens~\cite{soltau2016neural}. We tokenize the training data using a vocabulary of $66$K wordpieces~\cite{schuster2012japanese}. Thus, the input to the model is a sequence of wordpieces. 
Using wordpieces allows us to address out-of-vocabulary (OOV) word issues that would arise from having a fixed word vocabulary.
In practice, a wordpiece RNNLM gives similar performance as a word-level model~\cite{mielke2018spell}. For about $75\%$ of the segments, we were able to obtain visual features at the frame level. The features are $1500$-dimensional vectors, extracted from the video frames at $1$-second intervals, similar to those used for large scale image classification tasks \cite{varadarajan2015efficient,45619}. 
For a $K$-second video and $N>K$ wordpieces, each feature is uniformly allocated to $N/K$ wordpieces.

Our RNNLM models consist of 2 LSTM layers, each containing 2048 units which are linearly projected to 512 units \cite{sak2014}. The word-piece and video embeddings are of size $512$ each. We \emph{do not} use dropout. During training, the batch size per worker is set to 256, and we perform full length unrolling to a max length of 70. The $l^2$-norms of the gradients are clipped to a max norm of $1.0$ for the LSTM weights and to 10,000 for all other weights. We train with Synchronous SGD with the Adafactor optimizer \cite{shazeer2018adafactor} until convergence on a development set, created by randomly selecting $1\%$ of all utterances.

\section{Experiments}
\begin{table}[t]
    \centering
    \begin{tabular}{c|cc}
    \toprule
        Inputs to  & \multicolumn{2}{c}{Perplexity} \\
        \textit{Middle Fusion} & \youcook & \smg\\
    \midrule
        text + video & 64.9 & 411.4\\
        text + zero vectors & 99.0 & 537.7 \\
    \bottomrule
    \end{tabular}
    \caption{Withholding visual context from our best model leads to worse performance (similar to an RNNLM trained only on text).}
    \label{tab:middle}
\end{table}
\begin{table*}[t!]
\newcommand{\match}[1]{\hl{#1}}
    \centering
    \begin{tabular}{c|cccccccccc@{\hspace{5.2\tabcolsep}}c|c}
    
        & spray & the& pan& with& cooking& spray &&&&&& \textbf{Total Score} \\
        text-only & 10.2 & 2.9 & 5.6 & 1.4 & 7.5 & 0.2 &&&&&& 27.8 \\ 
        \textit{Middle Fusion} & \match{\textbf{8.9}} & 2.6 & \match{\textbf{3.5}} & 1.8 & 7.7 & 0.5 &&&&&& \textbf{24.8}\\
    \bottomrule
    \multicolumn{13}{c}{\small a) Our multimodal model has significantly lower word-level perplexity on}\\
    \multicolumn{13}{c}{\small word-pieces that correspond to items shown in the video (``spray, pan").}
    \end{tabular}
    
    \vspace*{2ex}
    \begin{tabular}{c|c@{\hspace{1\tabcolsep}}c@{\hspace{1\tabcolsep}}c@{\hspace{1\tabcolsep}}c@{\hspace{1\tabcolsep}}c@{\hspace{1\tabcolsep}}c@{\hspace{1\tabcolsep}}c@{\hspace{1\tabcolsep}}c@{\hspace{1\tabcolsep}}c@{\hspace{1\tabcolsep}}c@{\hspace{1\tabcolsep}}c|c}
        & place & cucumber & salad & and & then & the & hot & dog & on & the & bun & \textbf{Total Score}\\
        text-only& 7.7 & 14.5 & 4.8 & 1.9 & 2.9 & 3.3 & 6.6 & 3.1 & 4.5 & 0.9 & 5.0 & \textbf{55.1}\\
        \textit{Middle Fusion}  & 6.6 & \match{\textbf{11.9}} & 5.3 & 2.0 & 4.0 & 3.5 & 6.1 & 4.6 & 4.0 & 1.3 & 7.7 & 57.0\\
    \bottomrule
    \multicolumn{13}{c}{\small b) A rare example where the text-only model is overall better that the multimodal one. Still, though, }\\
    \multicolumn{13}{c}{\small entities (``cucumber") that appear in the video receive better scores from the multimodal system.}
    \end{tabular}
    \caption{Two sentences from \youcook{} with wordpiece-level negative log likelihood scores. Most gains (\match{highlighted}) of our \textit{Middle Fusion} model come from word-pieces corresponding to entities that appear in the videos.}
    \label{tab:examples}
\end{table*}
For evaluation we used two datasets, \youcook{} and \smg{}, allowing us to evaluate our models in cases where the visual context is relevant to the modelled language.
Note that no data from these datasets are present in the YouTube videos used for training. The perplexity of our models is shown in Table~\ref{tab:ppl}.

\paragraph{YouCookII dataset:} The YouCookII dataset (\youcook) \cite{das2013thousand,zhou2018towards} consists of 2,000 instructional cooking videos, each annotated with steps localized in video. An example annotation could be that of a video segment between 00:53--01:03 with the recipe step ``cook bacon until crispy, then drain on paper towel." The dataset was manually created, so that for each textual recipe segment the corresponding video frame provides related context. Therefore, this constrained scenario allows us to explicitly test whether our language models indeed manage to take advantage of the visual context.
    
\paragraph{20BN-Something-Something dataset v2: \cite{goyal2017something}:} This dataset (henceforth \smg) consists of about $220$K short videos of actions performed by humans with every day objects, annotated with text descriptions of the actions. Each description follows a template e.g. ``Taking something out of something."
This is a very constrained scenario, where the objects (``something") mentioned in the text definitely appear in the video.
We evaluate on the predefined validation set ($25$K videos) computing perplexity (PPL) on the textual action descriptions.\\

Out of all the architectures we consider, only two lead to consistently better performance on both datasets: \textit{Middle} and \textit{Late Fusion}. \textit{Late Fusion} leads to modest improvements (12\% and 5\% relative on the two datasets), but \textit{Middle Fusion} is able to take better advantage of both modalities, leading to 28\% and 20\% relative reductions in perplexity. In contrast, \textit{Early Fusion} performs worse than the baseline.
We suspect that the crucial factor for success with such architectures is allowing at least one lower layer of the RNNLM to be dedicated to text-only modeling. 

The variants that do not simply concatenate the word and video embeddings, but perform a fine-grainer integration, yield improvements on only one dataset (\youcook{}). The linear combination approach leads to 6\% relative reduction, while the learned weighting of the video embedding reduces perplexity by 14\%. 

The domain shift between training and the \smg{} dataset is reflected in quite high PPL scores. The videos are also much shorter (typically a few seconds) than the average YouTube video. We speculate that the length mismatch between training and test is responsible for the lower performance of the fine-grained approaches on \smg{}.

\paragraph{Does our model really use the visual features?}
In order to confirm that our model does utilize the visual modality, we perform a simple experiment of \emph{blinding} it: we deprive the RNNLM of the visual context, substituting the video embeddings with zero vectors.
The the results shown in Table~\ref{tab:middle}. 
The performance is worse, but it is in fact comparable to a model trained only on the text modality on \youcook. 
This confirms that our model indeed uses the visual context in a productive way.
Furthermore, it shows that our model is somewhat robust to the absence of visual context; this is the result of training with 25\% of our instances lacking visual features.

\paragraph{Where do the improvements come from?} We obtained wordpiece-level negative log likelihoods for 50 randomly chosen sentences from the \youcook{} dataset. For the majority (88\% of the sentences), the \textit{Middle Fusion} model had better sentence-level scores than the text-only model. We show two examples in Table~\ref{tab:examples}. We find that the largest improvements are due to the added visual information: the highest reductions are found on word-pieces corresponding to entities that appear in the video.

\section{Conclusion}
We present a simple strategy to augment a standard recurrent neural network language model with temporal visual features. Through an 
exploration of candidate architectures, we show that the \textit{Middle Fusion} of visual and textual features leads to a 20-28\% reduction in perplexity relative to a text only baseline.
These experiments were performed using datasets of unprecedented scale, with more than 1.2 billion tokens -- two orders of magnitude more than any previously published work. Our work is a first step towards creating and deploying large-scale multimodal systems that properly situate themselves into a given context, by taking full advantage of every available signal.


\bibliographystyle{acl_natbib}
\bibliography{references}

\end{document}